\documentclass[letterpaper, 10 pt, conference]{ieeeconf}  

\IEEEoverridecommandlockouts                              

\usepackage{graphicx} 
\usepackage{amsmath} 
\usepackage{amssymb}  
\usepackage{booktabs}    
\usepackage{multirow}    
\usepackage{makecell}
\usepackage{caption}
\usepackage{cuted}
\usepackage{hyperref}

\title{\LARGE \bf
ReMAP-DP: Reprojected Multi-view Aligned PointMaps for Diffusion Policy
}

\author{Xinzhang Yang$^{1\ast}$, Renjun Wu$^{1\ast}$, Jinyan Liu$^{1}$ and Xuesong Li$^{1 \dagger}$
\thanks{$^{\ast}$ Equal contribution.}
\thanks{$^{\dagger}$ The corresponding author:{ \tt\small lixuesong@bit.edu.cn}}
\thanks{$^{1}$ School of Computer Science, Beijing Institute of Technology, China.}%
}

\begin{document}

\maketitle

\thispagestyle{empty}
\pagestyle{empty}

\begin{strip}
    \vspace{-1cm}
    \centering
    \includegraphics[width=\textwidth]{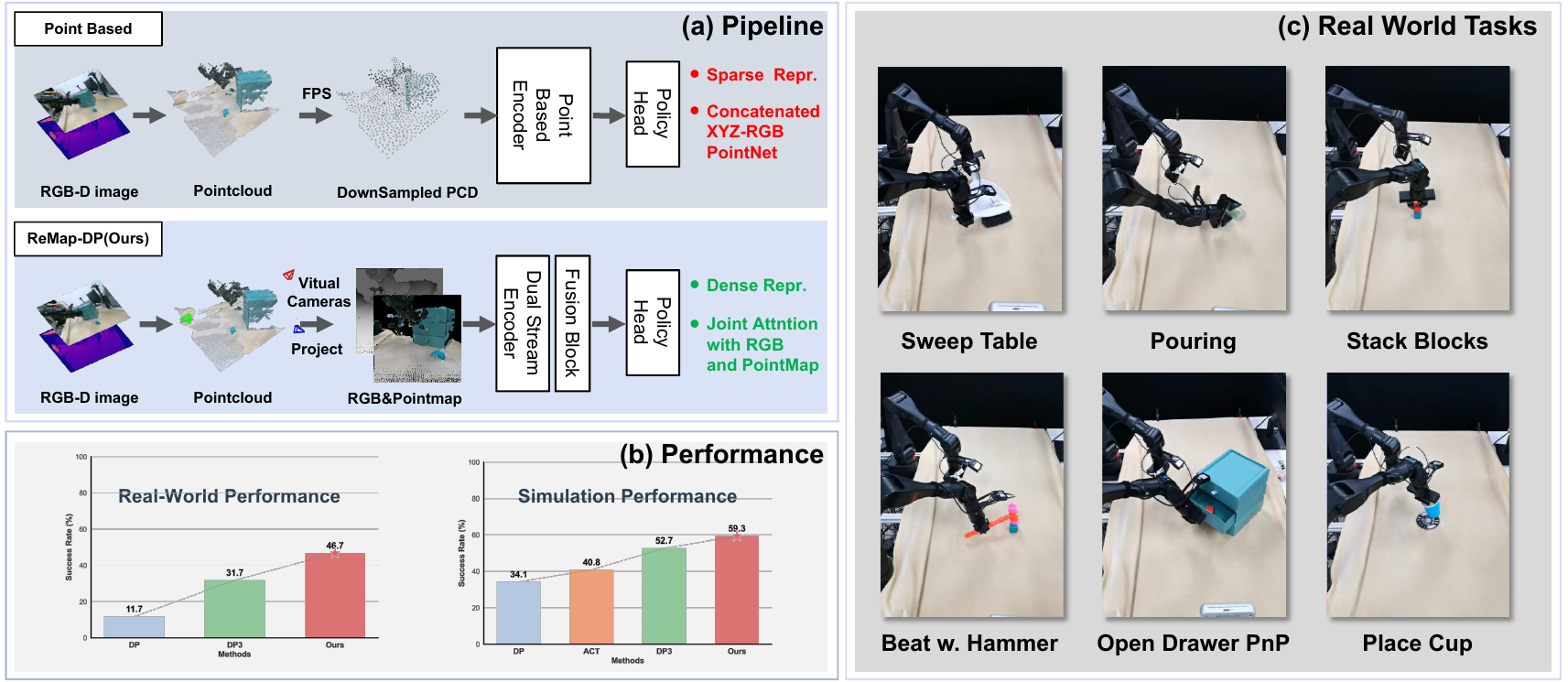}
    \captionof{figure}{\textbf{(a) ReMAP-DP} is a multi-view projection framework for visuomotor control. Unlike approaches relying on sparse point cloud downsampling, our work leverages dense PointMaps and RGB inputs, synergizing explicit geometric structures with visual semantics through cross-modal transformer fusion. \textbf{(b)} The experimental results indicate that ReMAP-DP outperforms multiple strong baselines and \textbf{(c)} demonstrates robust performance across a diverse range of real-world tasks.}
    \label{fig:title}
\end{strip}

\IEEEpeerreviewmaketitle

\begin{abstract}

Generalist robot policies built upon 2D visual representations excel at semantic reasoning but inherently lack the explicit 3D spatial awareness required for high-precision tasks. Existing 3D integration methods struggle to bridge this gap due to the structural irregularity of sparse point clouds and the geometric distortion introduced by multi-view orthographic rendering. To overcome these barriers, we present ReMAP-DP, a novel framework synergizing standardized perspective reprojection with a structure-aware dual-stream diffusion policy. By coupling the re-projected views with pixel-aligned PointMaps, our dual-stream architecture leverages learnable modality embeddings to fuse frozen semantic features and explicit geometric descriptors, ensuring precise implicit patch-level alignment. Extensive experiments across simulation and real-world environments demonstrate ReMAP-DP's superior performance in diverse manipulation tasks. On RoboTwin 2.0, it attains a 59.3\% average success rate, outperforming the DP3 baseline by +6.6\%. On ManiSkill 3, our method yields a 28\% improvement over DP3 on the geometrically challenging Stack Cube task. Furthermore, ReMAP-DP exhibits remarkable real-world robustness, executing high-precision and dynamic manipulations with superior data efficiency from only a handful of demonstrations. Project page is available at: \url{https://icr-lab.github.io/ReMAP-DP/}

\end{abstract}

\section{INTRODUCTION}

Developing generalist robotic agents capable of diverse manipulation in unstructured environments remains a fundamental goal \cite{ours:OpenVLA}. While Vision-Language-Action (VLA) models \cite{ours:Pi0,ours:RT2,ours:Octo} and diffusion-based visuomotor policies \cite{ours:ACT,ours:DP} demonstrate remarkable capabilities by leveraging 2D RGB images and pre-trained vision foundation models \cite{ours:DINOV2,ours:CLIP,ours:SigLIP}, they inherently lack explicit 3D geometric scene information. Consequently, this spatial deficit severely hinders performance in tasks demanding precise depth perception and fine-grained actuation \cite{ours:SpatialVLA,ours:PointVLA}.

Prior work addressing this spatial deficit faces distinct architectural trade-offs. Implicit feature injection \cite{ours:VGGTDP,ours:SpatialForcing} suffers from severe domain gaps and computationally prohibitive fine-tuning. Explicit 3D representations provide direct grounding, yet voxel methods \cite{ours:PerAct} incur cubic computational costs, and sparse point clouds \cite{ours:DP3} possess structural irregularity that resists standard vision architectures. Attempting to resolve this irregularity, PointMapPolicy \cite{ours:PointMapPolicy} structures 3D coordinates into dense 2D matrices. However, single-camera extraction tightly overfits the policy to specific hardware. While multi-view virtual rendering \cite{ours:RVT,ours:RVT2} circumvents view-dependency, its strict reliance on orthographic projection distorts natural perspective convergence, catastrophically disrupting the visual priors of pre-trained 2D foundation models.

Beyond these representational limitations lies a more fundamental bottleneck: effectively fusing high-dimensional semantic priors with explicit geometric structures. Existing methods persistently treat RGB and depth modalities as loosely coupled inputs via shallow channel concatenation. Such naive integration fails to explicitly align semantic identity with spatial occupancy at the token level, causing policies to struggle with high-precision tasks requiring an exact, correlated understanding of \textit{what} an object is with \textit{exactly where} it is located.

To overcome these interrelated challenges, we present ReMAP-DP (\textbf{Re}projected \textbf{M}ulti-view \textbf{A}ligned \textbf{P}ointMaps for Diffusion Policy). Our pipeline initiates with a \textbf{Workspace Multi-View Projection} module that lifts synchronized RGB-D inputs into a unified point cloud, re-projecting it to generate aligned perspective RGB images and explicit PointMaps. Establishing this canonical observation space harmonizes heterogeneous camera configurations while preserving natural perspective geometry. To encode these inputs, we propose a \textbf{Dual-Stream Observation Encoding} architecture: a semantic stream uses a frozen DINOv2 \cite{ours:DINOV2} to extract spatial patch tokens, while a parallel geometric stream processes the PointMap via a trainable Projective PointMap Vision Transformer (ViT) \cite{ours:ViT} to extract precise geometric tokens. Crucially, to resolve the fusion bottleneck, we introduce a \textbf{Multi-Modal Transformer Fusion} module. Since both token sets originate from identical 2D grids, learnable modality embeddings explicitly guide the cross-modal transformer to distinguish semantic texture from spatial structure, yielding a highly correlated global observation embedding. Finally, for \textbf{Conditional Diffusion Action Generation}, this embedding conditions a 1D Temporal U-Net Denoising Diffusion Probabilistic Model (DDPM) \cite{ours:DDPM} via FiLM \cite{ours:FiLM} layers to iteratively reconstruct precise robot action trajectories.

Extensive evaluations on ManiSkill3 \cite{ours:maniskill3} and RoboTwin 2.0 \cite{ours:robotwin2} demonstrate that ReMAP-DP exhibits superior performance against leading visuomotor policies, notably outperforming the strong DP3 baseline. Furthermore, real-world deployments show ReMAP-DP  generalizes to high-precision tasks using only a few dozen demonstrations.

In summary, our contributions are as follows:
\begin{itemize}
\item We propose a multi-view perspective reprojection framework that unifies heterogeneous camera inputs into a canonical observation space, facilitating cross-dataset generalization while preserving the natural perspective appearance essential for 2D foundation models.
\item We introduce a structure-aware dual-stream fusion architecture. By employing learnable modality embeddings within a cross-modal transformer, we achieve implicit patch-level alignment between frozen semantic priors and explicit geometry features.
\item We demonstrate superior performance in both extensive simulations and real-world robot experiments. Comprehensive ablation studies validate our underlying structural representations and multi-modal fusion architecture.
\end{itemize}

\section{RELATED WORKS}

\subsection{Multi-View Representation}
While multi-view representations excel in broader computer vision tasks \cite{ours:RL1_1_50,ours:RL1_1_53,ours:RL1_4_21}, robotic frameworks like Robot Vision Transformer (RVT) \cite{ours:RVT} adapt this paradigm via virtual multi-view rendering to effectively decouple policies from physical camera extrinsics. However, such methods typically rely on orthographic projection—distorting natural perspective convergence—fuse modalities naively at the channel level, and restrict action to key-frame planning \cite{ours:RT2,ours:PerAct}. Resolving these limitations, we advances multi-view policies by employing standardized perspective reprojection to preserve visual priors, while utilizing a structure-aware dual-stream architecture to explicitly align modalities for continuous action diffusion.

\subsection{3D Representations for Robotic Manipulation}
End-to-end 2D visuomotor policies \cite{ours:RT1,ours:Octo,ours:OpenVLA} leverage robust foundation model semantics but require extensive demonstrations to implicitly infer scene geometry \cite{ours:RT2}. Seeking explicit geometric awareness, early approaches utilized voxelized workspaces \cite{ours:PerAct,ours:C2F} or NeRF-based renderings \cite{ours:RL2_4_9,ours:GNFactor}, yet cubic costs and degraded quality severely restrict their scalability. As an efficient alternative, sparse point cloud paradigms like 3D Diffusion Policy (DP3) \cite{ours:DP3} enhance spatial generalization. Nevertheless, encoding unordered points via simple MLPs or Farthest Point Sampling inherently discards the dense local geometry critical for high-precision manipulation \cite{ours:3DDA}.

Attempts to fuse 3D modalities into 2D foundation models face significant hurdles. Embedding features directly into the visual space \cite{ours:3DVLA,ours:SpatialVLA} disrupts pre-trained alignments, demanding costly retraining. Alternatively, parallel encoders \cite{ours:3DDA,ours:FPVNet} struggle to map dense 2D semantic grids onto irregular point clouds. ReMap circumvents these bottlenecks by extracting 3D geometry as grid-aligned PointMaps. This structural isomorphism enables precise patch-level integration of spatial features with frozen semantic priors, fully preserving the foundation model's integrity.

\subsection{Diffusion Models} 
Beyond foundational success in image generation \cite{ours:DDPM,ours:SDE,ours:rl_3_2_51}, diffusion models now drive advancements in reinforcement learning \cite{ours:rl3_2_70}, motion planning \cite{ours:rl3_4_18}, and large-scale generalist policies \cite{ours:Octo,ours:RDT1B}. Moving beyond visual subgoal prediction \cite{ours:dallebot}, recent imitation learning paradigms formulate visuomotor policies directly as conditional diffusion models for continuous action generation \cite{ours:DP}. However, their efficacy is fundamentally bottlenecked by conditioning signal quality. Prior 3D-aware policies rely on raw scene features \cite{ours:rl3_3_21} or 1D point cloud embeddings \cite{ours:DP3}, introducing structural irregularities that hinder fine-grained multi-modal alignment. Overcoming this, we condition an efficient 1D diffusion policy \cite{ours:DP} on structure-aligned PointMaps, resolving input irregularity to enable precise 6-DoF trajectory generation.

\begin{figure*}[t]
    \vspace{8pt}
    \centering
    \includegraphics[width=\textwidth]{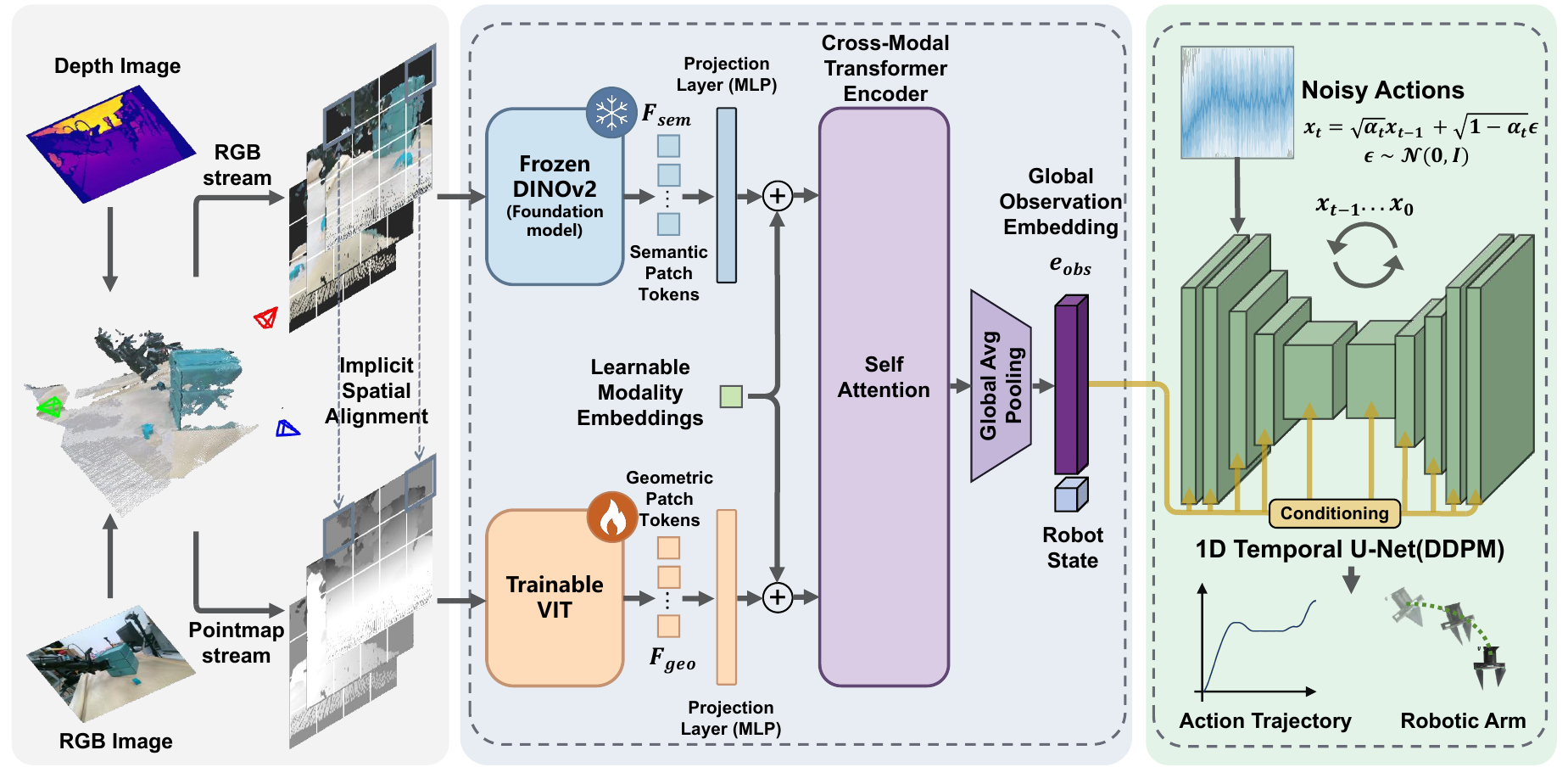}
    \caption{ \textbf{Overall architecture of ReMAP-DP}. The method consists of three components: (1) Back projection and re-projection are used to obtain RGB and point maps for workspace alignment and denoising. (2) 
A dual stream encoder and transformer is used to process RGB and geometric features and fuse them together. (3) An action generation module employs a 1D-UNet Diffusion Policy to predict precise robot actions conditioned on the fused multi-modal features.}
    \label{fig:main method}
\end{figure*}

\section{METHOD}

In this section, we present our framework for high-precision robotic manipulation. We formulate the control problem as learning a conditional distribution over action trajectories via Denoising Diffusion Probabilistic Models (DDPM) \cite{ours:DDPM}. To address the challenge of aligning semantic understanding with precise geometry in cluttered environments, we introduce a \textbf{Dual-Stream Spatio-Semantic Architecture}. By processing visual inputs through parallel semantic and geometric streams, this design enables exact patch-level fusion via a cross-modal transformer, yielding a structurally aligned multimodal embedding to condition the action diffusion process.

\subsection{Problem Formulation}
We consider a visuomotor policy learning setting where the robot observes the environment through RGB-D sensors. Let $O_t$ denote the observation at time step $t$, comprising an RGB image $I_{rgb}$, an aligned depth map $I_{depth}$, and proprioceptive robot states $q_t$. Let $\mathbf{a}_t$ denote the robot action, consisting of the joint positions and gripper status.

Instead of predicting a single-step action, we adopt the Diffusion Policy formulation \cite{ours:DP} to predict a sequence of future actions $A_t = [\mathbf{a}_t, \dots, \mathbf{a}_{t+H}]$, where $H$ is the prediction horizon. The policy $\pi_\theta(A_t | o_t)$ is trained to approximate the conditional distribution of demonstrated trajectories by iteratively denoising a Gaussian noise variable $A^K_t \sim \mathcal{N}(0, \mathbf{I})$ into the action sequence $A^0_t$ via $K$ steps.

\subsection{Dual-Stream Observation Encoding}

To enable robust manipulation, the robot must perceive both \textit{what} objects are (semantics) and \textit{where} they are located spatially (geometry). We propose a \textbf{Dual-Stream Observation Encoder} to address this challenge. The core of our perception module is a synchronized dual-stream pipeline that processes re-projected visual data.

\subsubsection{Workspace Multi-View Projection}
The standard manipulation strategies trained on raw RGB-D data often suffer from viewpoint bias, where the camera angle relative to the workspace varies between data collection and deployment. To unify visual observations into a consistent representation, we aggregate raw RGB-D inputs into a global colored 3D point cloud $P_{rgbxyz}$. We subsequently project this point cloud via multi-view perspective projection to generate pixel-aligned \textbf{re-projected RGB images} ($I_{rgb}$) and dense \textbf{coordinate pointMaps} ($I_{geo}$).

\textbf{Canonical View Invariance.} A critical advantage of this projection pipeline is the decoupling of the observation space from the physical sensor pose. By rendering the point cloud from fixed, virtual canonical viewpoints, which are empirically configured relative to the robot base to optimally cover the manipulation workspace, we ensure that the policy inputs remain consistent even if the physical cameras are perturbed or if the robot operates in a dynamic setting.
Furthermore, by design, this projection guarantees a strict pixel-wise alignment where every pixel $(u,v)$ in $I_{rgb}$ maps one-to-one to the physical metric coordinate stored at $(u,v)$ in $I_{geo}$. This establishes an implicit yet rigorous spatial correspondence between semantic appearance and geometric structure, eliminating the need for the network to learn alignment from scratch.

\subsubsection{Semantic Stream: Spatial-Aware Foundation Model}

To extract robust visual features resilient to lighting and texture variations, we leverage the DINOv2 (ViT-S/14) \cite{ours:DINOV2} foundation model. Unlike supervised baselines (e.g., ImageNet-ResNet) which prioritize global class separability, the self-supervised objective of DINOv2 encourages the emergence of localized, part-aware features without explicit labels.
Standard usage often involves collapsing features into a global \textbf{[CLS]} token; however, precise manipulation requires retaining high-fidelity spatial localization. Therefore, we discard the global token and explicitly extract the patch-level features to preserve the spatial topology of the scene.
Given the re-projected RGB input $I_{rgb}$, the encoder outputs a sequence of location-aware tokens:
$$
F_{rgb} = \text{DINOv2}(I_{rgb}) \in \mathbb{R}^{N \times D_{sem}}
$$
where $N=256$ represents the number of patches and $D_{sem}=384$ is the semantic embedding dimension.
To adapt these features to our specific manipulation tasks without disrupting the robust pre-trained manifold, we freeze the backbone weights. We utilize a lightweight linear projection $\phi_{rgb}$ to map these tokens to the shared fusion dimension $D_{fusion}$. This linear mapping preserves the geometric relationships learned by the foundation model while aligning the subspace with the geometric stream.

\subsubsection{Geometric Stream: Projective PointMap ViT}

Complementing the semantic stream, the geometric stream processes the aligned PointMap $I_{geo}$. Unlike a standard depth map which only encodes relative distance $z$, our PointMap explicitly encodes absolute metric coordinates $(x, y, z)$ in its three channels. This ensures that the network has direct access to the workspace coordinates required for converting visual features into 6-DoF action poses.

To ensure structural symmetry with the semantic backbone, we employ a custom Vision Transformer (ViT) trained from scratch. Crucially, we configure this encoder with a patch size ($P=14$) identical to the DINOv2 backbone, ensuring that the resulting token grid spatially aligns with the semantic tokens.
The PointMap is patchified, flattened, and projected into an embedding space before processing:
$$
F_{geo} = \text{ViT}_{pmp}(I_{geo}) \in \mathbb{R}^{N \times D_{geo}}
$$
Here, the encoder consists of a 4-layer Transformer with an embedding dimension of $D_{geo}=128$ Similar to the semantic stream, these geometric tokens are linearly projected to $D_{fusion}$, creating a pair of structurally aligned feature sequences ready for cross-modal fusion.

\subsubsection{Multi-Modal Transformer Fusion}

A naive concatenation of multi-modal features fails to explicitly capture the interplay between semantics and geometry. We address this bottleneck via a joint attention mechanism designed to exploit the structural symmetry of the parallel encoding streams.

\textbf{Implicit Spatial Alignment.} A critical design choice is the synchronization of patch sizes ($P=14$) and input resolutions ($224 \times 224$) across both the semantic and geometric encoders. Consequently, the $i$-th token in $F_{rgb}$ corresponds to the exact identical physical workspace region as the $i$-th token in $F_{geo}$. This \textit{implicit spatial alignment} significantly reduces the learning burden, allowing the fusion module to focus on feature correlation rather than learning spatial correspondences from scratch.

\textbf{Modality-Aware Joint Attention.} To fuse these modalities, we first project both token sequences to the shared dimension $D_{fusion}$. However, since both streams are projected to the same latent space, the self-attention mechanism requires explicit signals to distinguish visual texture information from geometric coordinate information. Therefore, We introduce learnable \textbf{modality embeddings} $E_{mod} \in \mathbb{R}^{D_{fusion}}$ to the respective streams:
$$
S = \left[ \phi_{rgb}(F_{rgb}) + E_{mod}^{rgb}, \quad \phi_{geo}(F_{geo}) + E_{mod}^{geo} \right]
$$
The resulting sequence $S \in \mathbb{R}^{2N \times D_{fusion}}$ is processed by a 2-layer transformer encoder with self-attention. This mechanism allows the model to dynamically weigh geometric details conditioned on semantic identity. Finally, we apply Global Average Pooling (GAP) \cite{ours:GAP} to obtain the compact observation embedding $z_{obs} \in \mathbb{R}^{D_{fusion}}$.

\begin{figure*}[t]
    \vspace{8pt}
    \centering
    \includegraphics[width=\textwidth]{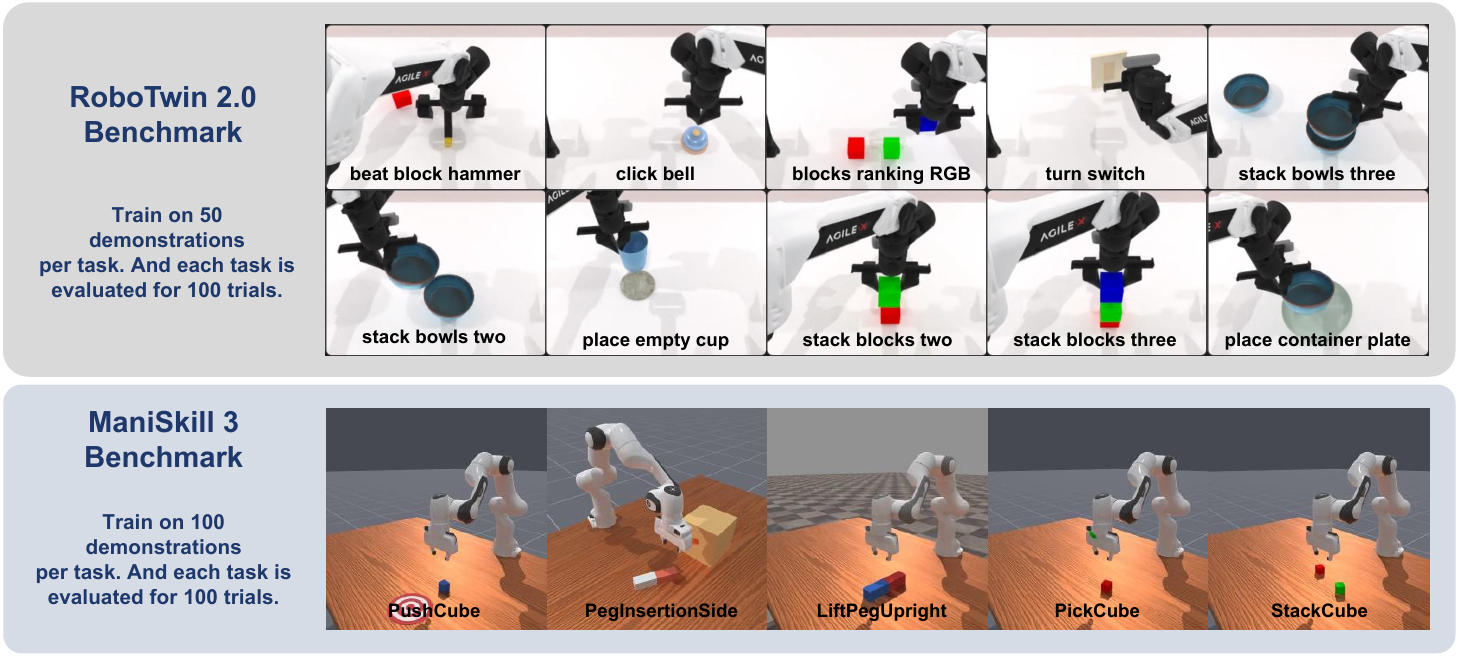}
    \caption{\textbf{Top: }Visualization of ten selected tasks in RoboTwin 2.0 Benchmarks. \textbf{Bottom: }Visualization of five selected tasks in ManiSkill 3 Benchmarks.}
    \label{fig:simulation}
\end{figure*}

\subsection{Conditional Diffusion Action Generation}

The action generation process of the policy is modeled using a Conditional Denoising Diffusion Probabilistic Model (DDPM). We utilize a 1D Temporal U-Net $\epsilon_\theta$ as the noise prediction network, which takes the noisy action sequence $A_k$, the diffusion timestep $k$, and the global observation condition $z_{obs}$ as inputs.

\textbf{U-Net Architecture And Conditioning.}
The U-Net consists of a series of down-sampling and up-sampling purely convolutional blocks. Unlike standard cross-attention conditioning, we utilize Feature-wise Linear Modulation (FiLM) \cite{ours:FiLM} to inject the global observation $z_{obs}$ into the network. For a feature map $h$ at any layer of the U-Net, FiLM applies an affine transformation predicted from the condition:
$$
\text{FiLM}(h | z_{obs}) = \gamma(z_{obs}) \odot h + \beta(z_{obs})
$$
where $\gamma$ and $\beta$ are scale and shift parameters learned by a dense network. This method allows the global spatio-semantic context to modulate the action generation process at every level of abstraction, from coarse trajectory planning to fine motor control.

\textbf{Training Objective.}
We employ the standard DDPM training objective. During training, we sample a random timestep $k$, Gaussian noise $\epsilon \sim \mathcal{N}(0, \mathbf{I})$, and a ground-truth trajectory $A_0$ from the expert dataset. The network is optimized to predict the noise residual via the Mean Squared Error (MSE) loss:
$$
\mathcal{L} = \mathbb{E}_{k, \epsilon, A_0} \left[ \| \epsilon - \epsilon_\theta(\sqrt{\bar{\alpha}_k} A_0 + \sqrt{1 - \bar{\alpha}_k} \epsilon, k, z_{obs}) \|^2 \right]
$$
During inference, we adopt a squared-cosine beta schedule to control the noise levels $\bar{\alpha}_k$, ensuring high sample quality for precise manipulation tasks.

\begin{table*}[t]
\centering
\caption{Comparison of different methods across various manipulation tasks. Our method achieves the best average performance.}
\label{tab:manipulation_robotwin}
\setlength{\tabcolsep}{4pt} 
\begin{tabular*}{\textwidth}{l @{\extracolsep{\fill}} c c c c c c c c c c | c} 
\toprule
\textbf{Method} & 
\makecell{{Beat}\\{Block}\\{Hammer}} & 
\makecell{{Blocks}\\{Ranking}\\{RGB}} & 
\makecell{{Click}\\{Bell}} & 
\makecell{{Turn}\\{Switch}} & 
\makecell{{Place}\\{Container}\\{Plate}} & 
\makecell{{Place}\\{Empty}\\{Cup}} & 
\makecell{{Stack}\\{Blocks}\\{Two}} & 
\makecell{{Stack}\\{Blocks}\\{Three}} & 
\makecell{{Stack}\\{Bowls}\\{Three}} & 
\makecell{{Stack}\\{Bowls}\\{Two}} & 
\textbf{Avg.} \\
\midrule
\textbf{Ours} & 0.70 & \textbf{0.03} & \textbf{1.00} & \textbf{0.60} & 0.80 & \textbf{0.68} & \textbf{0.50} & \textbf{0.08} & \textbf{0.72} & 0.82 & \textbf{0.593} \\
DP3 \cite{ours:DP3} & \textbf{0.72} & \textbf{0.03} & 0.90 & 0.46 & \textbf{0.86} & 0.65 & 0.24 & 0.01 & 0.57 & \textbf{0.83} & 0.527 \\
ACT \cite{ours:ACT} & 0.56 & 0.01 & 0.58 & 0.05 & 0.72 & 0.61 & 0.25 & 0.00 & 0.48 & 0.82 & 0.408 \\
DP \cite{ours:DP} & 0.42 & 0.00 & 0.54 & 0.36 & 0.41 & 0.37 & 0.07 & 0.00 & 0.63 & 0.61 & 0.341 \\
\bottomrule
\end{tabular*}
\end{table*}

\begin{figure*}[t]
    \centering
    \includegraphics[width=\textwidth]{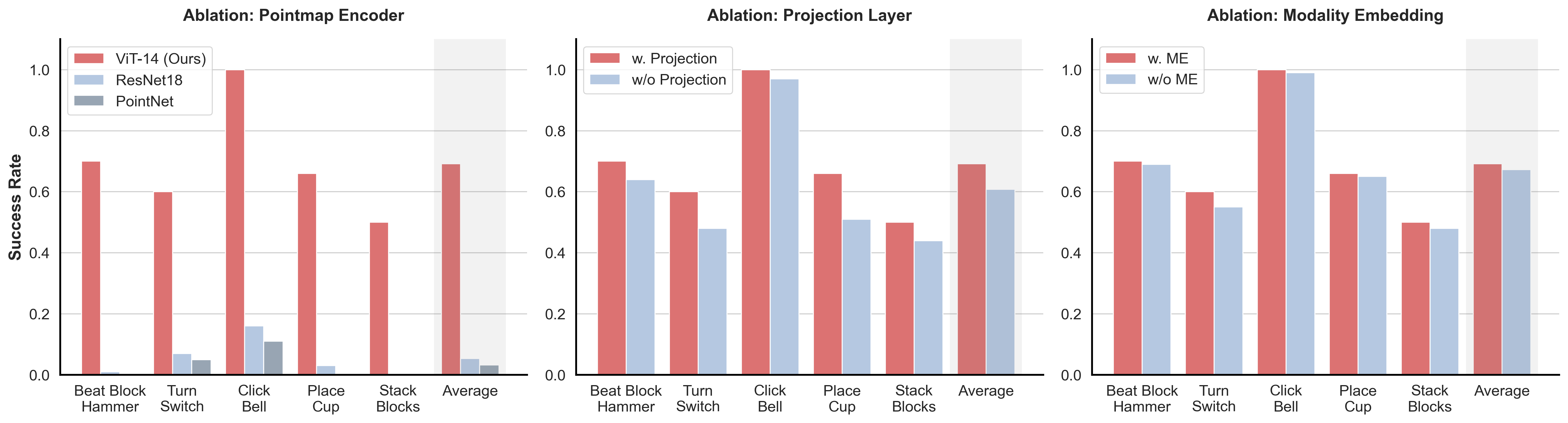}
    \caption{\textbf{Left:} Architecture of Geometry Encoder. \textbf{Mid: }Efficacy of Multi-View Projection. \textbf{Right: }Impact of Modality Embeddings.}
    \label{fig:ablation}
\end{figure*}

\section{EXPERIMENTS}

In this section, we evaluate the proposed framework in both simulated and real-world environments. We design experiments to answer the following research questions:
\begin{itemize}
    \item \textbf{RQ1 (Performance):}  Does the proposed dual-stream spatio-semantic fusion outperform state-of-the-art baselines in high-precision manipulation tasks?
    \item \textbf{RQ2 (Ablation):}  What is the contribution of each component in our architecture design, including the spatial reprojection strategy, the geometric encoder backbone, and the explicitly learned modality embeddings?
\end{itemize}   

\subsection{Experimental Setup}

\subsubsection{Simulation Benchmarks} We utilize two diverse benchmarks. \textbf{ManiSkill 3} evaluates generalizable skills using a Franka Emika Panda, for which we collect 100 expert demonstrations per task. \textbf{RoboTwin 2.0} supplies a variety of dual arm desktop tasks; here We select 10 tasks in the "Easy" setting with 50 demonstrations each. For robust evaluation, we conduct 100 trials per checkpoint.

\subsubsection{Implementation Details}

We implement our framework using PyTorch and train all models on a single workstation equipped with an NVIDIA RTX 4090.

\textbf{Network Architecture:}
The perception module processes $224 \times 224$ inputs.

\begin{itemize}
    \item \textbf{Semantic Stream:} We employ a DINOv2-ViT-S/14 encoder. This backbone is frozen during training. The output feature dimension is 384. A linear projection layer maps these features to the shared fusion dimension ($D_{fusion} = 256$).
    \item \textbf{Geometric Stream:} The PointMap encoder utilizes a custom Vision Transformer (ViT) architecture trained from scratch. Key parameters include: input channels $C=3$, patch size $P=14$, embedding dimension $D_{embed}=128$, depth $L=4$, and number of attention heads $H=4$. A separate linear projection layer maps the 128-dimensional geometric features to the shared dimension $D_{fusion}=256$.
    \item \textbf{Fusion Module:} The cross-modal fusion is handled by a 2-layer Transformer Encoder. We configure it with a model dimension $d_{model}=256$, 4 attention heads, a feedforward dimension of 1024, dropout of 0.1, and GeLU activation.
\end{itemize}

\textbf{Optimization:} 
We use the AdamW optimizer with a learning rate of $1.0 \times 10^{-4}$ and weight decay of $1.0 \times 10^{-6}$. The diffusion denoising process utilizes a squared-cosine noise schedule over $K=100$ inference steps.

\subsubsection{Baselines}
We compare our method against the following strong imitation learning approaches:
\begin{itemize}
    \item \textbf{Diffusion Policy (CNN-based):}  The standard image-to-action diffusion baseline using ResNet-18 encoders, which is a strong baselien for 2D visual imitation.
    \item \textbf{3D Diffusion Policy (DP3):}  A point-cloud-based diffusion policy using a PointNet++ encoder. This strong baseline represents pure geometric approaches without pre-trained semantic understanding.
    \item \textbf{ACT:}  Action Chunking with Transformers, a VAE-based policy utilizing separate modalities. 
\end{itemize}

\subsection{Simulation Results}

We report the success rates of our proposed method compared to the baseline approaches on both the RoboTwin 2.0 and ManiSkill 3 benchmarks. All results are averaged over 100 evaluation episodes per checkpoint.

\subsubsection{Performance on RoboTwin 2.0 Benchmark}

Table I summarizes the results on the RoboTwin 2.0 benchmark across 10 diverse manipulation tasks. Our method achieves the highest average success rate of \textbf{59.3\%}, significantly outperforming the standard UNet-based Diffusion Policy (34.1\%) and the transformer-based ACT (40.8\%).

Notably, our approach demonstrates superior performance in tasks requiring precise spatial reasoning and multi-stage planning, such as \textit{Stack Blocks Two} (50\% vs. 24\% for DP3) and \textit{Stack Bowls Three} (72\% vs. 57\%). For rigid object interactions like \textit{Click Bell} and \textit{Turn Switch}, our method achieves success rates of 100\% and 60\% respectively, surpassing the 3D-geometric baseline (DP3) by large margins. While DP3 performs marginally better on specific pick-and-place tasks like \textit{Place Container Plate} (86\% vs. 80\%), our framework maintains competitive performance while offering far stronger generalization on semantic-heavy tasks where pure geometric approaches struggle.

\begin{table}[h]
\centering
\caption{Success Rates on Maniskill 3 Benchmark}
\label{tab:manipulation_maniskill}
\begin{tabular*}{\columnwidth}{l @{\extracolsep{\fill}} c c c c c | c}
\toprule
\textbf{Method} & 
\makecell{{Push}\\{Cube}} &
\makecell{{Stack}\\{Cube}} &
\makecell{{Pick}\\{Cube}} &
\makecell{{Lift}\\{Peg}\\{Upright}} &
\makecell{{Peg}\\{Insertion}\\{Side}} &
\textbf{Avg.} \\
\midrule
\textbf{Ours} & \textbf{0.95} & \textbf{0.28} & \textbf{0.48} & 0.36 & 0.01 & \textbf{0.416}\\
DP3 & 0.92 & 0.01 & 0.24 & \textbf{0.75} & 0.01 & 0.386\\
DP & 0.88 & 0.00 & 0.44 & 0.03 & 0.01 & 0.272\\
\bottomrule
\end{tabular*}
\end{table}

\subsubsection{Performance on ManiSkill 3 Benchmark}
The higher degree of randomization in ManiSkill 3 yields lower overall success rates, as presented in Table \ref{tab:manipulation_maniskill}. Nevertheless, our method demonstrates competitive capabilities on fundamental manipulation skills. In the \textit{StackCube} task, which demands precise spatial alignment, our approach achieves a 28\% success rate. This significantly outperforms standard Diffusion Policy (0\%) and DP3 (1\%), demonstrating the efficacy of our dual-stream geometric alignment. Similarly, in \textit{PickCube}, it reaches 48\%, surpassing both DP3 (24\%) and DP (44\%). While DP3 retains an advantage on \textit{LiftPegUpright}, this likely reflects raw point clouds' inherent suitability for capturing local structural proportions rather than strict spatial alignment. 

It is worth noting that for the highly constrained \textit{PegInsertion} task, all methods struggle ($\le$1\%), suggesting a need for larger or higher-quality demonstration datasets. Overall, across the solvable tasks, our fused architecture provides the most stable policy.

\subsection{Ablation Study}

We validate our architectural choices using 5 RoboTwin 2.0 tasks (see Fig. \ref{fig:ablation}):

\begin{itemize}
    \item \textbf{Geometry Encoder:} We compare our dense PointMap ViT against sparse point-based (PointNet) and local convolution-based (ResNet-18) encoders to evaluate the efficacy of dense spatial encoding. The dense ViT yields a 69\% success rate, whereas sparse point sampling ($<5\%$) proves insufficient for tasks requiring strict metric alignment.
    
    \item \textbf{Multi-View Projection:} We compare feeding raw unaligned feature maps versus our re-projected inputs. Removing projection causes an 8\% performance drop—most notably in alignment-heavy tasks like \textit{Place Empty Cup} (-15\%)—confirming that explicit pixel-wise correspondence noticeably aids cross-modal learning.
    
    \item \textbf{Modality Embeddings:} We test fusion with and without explicit modality tags. Removing embeddings causes a minor degradation ($\sim$2\%), suggesting that while the network implicitly distinguishes inputs, explicit tagging provides a consistent optimization benefit for cross-modal fusion.
\end{itemize}

\subsection{Real-World Experiments}

To validate the effectiveness of our method in physical environments with sensor noise and lighting variations, we deployed the policy on a dual-arm robot platform.

\subsubsection{Hardware Setup}

\textbf{Robotic Platform:}  We utilize the Discover Airbot, a dual-arm system where each arm possesses 6 degrees of freedom (DoF) equipped with a parallel gripper. The robot is controlled via a high-frequency interface running on an NVIDIA RTX 4090 workstation.

\textbf{Perception System:}  The visual sensing array consists of three external RGB-D cameras: two Intel RealSense D455 units and one RealSense D435i unit, positioned at the front, side, and back of the workspace respectively to minimize occlusion. All cameras capture streams at a resolution of $640 \times 480$ at 30 FPS. We employ extrinsic calibration and ICP alignment to merge multi-view point clouds into a unified, high-fidelity geometry for projection. 

\begin{figure}[h]
    \centering
    \includegraphics[width=\columnwidth]{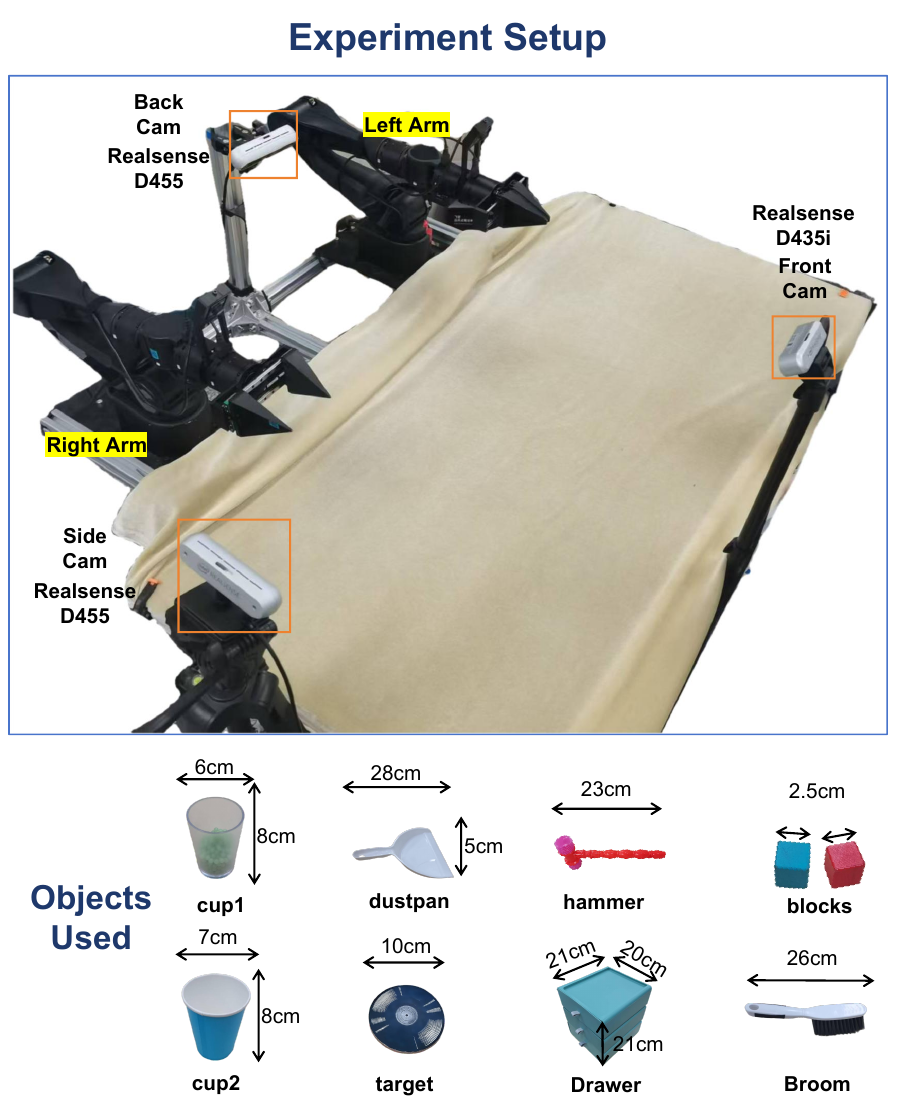}
    \caption{\textbf{Real-World Experiments}. \textbf{Top:} Real-World experiments setup, including cameras' positions and robotic arms' settings. \textbf{Bottom:} Objects used in tasks and their sizes.}
    \label{fig:realworld_setup}
\end{figure}

\subsubsection{Data Collection and Tasks}
We collected expert demonstrations using the Discover remote teleoperation platform. We designed 6 disparate tasks covering pick-and-place, dynamic manipulation (pouring), and articulated object interaction (drawer opening). For each task, we collected 50 demonstrations and executed 10 evaluation trials. To evaluate robustness, we applied randomized local perturbations to the initial positions and orientations of all movable objects prior to each trial.
\begin{table}[h]
\centering
\small
\setlength{\tabcolsep}{5pt}
\caption{Detailed descriptions for real world tasks.}
\label{tab:task_details_left}
\begin{tabular}{l p{5.5cm}}
\toprule
\textbf{Task} & \textbf{Task Details} \\
\midrule
\textit{Beat w. Hammer}  & Strike target object with hammer head \\
\textit{Sweep Table} & Sweep all rubbish into dustpan \\
\textit{Open Drawer PnP} & Open drawer, insert cube, close drawer \\
\textit{Place Empty Cup} & Stable placement in target area \\
\textit{Pouring} & Transfer small balls without spilling \\
\textit{Stack Cube} & Stable stack of red cube on blue cube \\
\bottomrule
\end{tabular}
\end{table}

\subsubsection{Results Analysis} 
Consistent with our simulation evaluations, we compare our real-world deployment against the DP and DP3 baselines. The quantitative results are reported in Table \ref{tab:task_details_right}. 
\begin{table}[h]
\centering
\small
\setlength{\tabcolsep}{5pt}
\caption{Success rates (SR) across 10 trials for different methods.}
\label{tab:task_details_right}
\begin{tabular*}{\columnwidth}{l @{\extracolsep{\fill}} c c c}
\toprule
\textbf{Task} & \textbf{Ours} & \textbf{DP3} & \textbf{DP} \\
\midrule
\textit{Beat w. Hammer} & \textbf{0.5} & 0.4& 0.2  \\
\textit{Sweep Table} & \textbf{0.5} & 0.3 &  0.1\\
\textit{Open Drawer PnP} & \textbf{0.3} & 0.1 & 0.0 \\
\textit{Place Empty Cup} & \textbf{0.6} & \textbf{0.6} & 0.3 \\
\textit{Pouring} & \textbf{0.6} & 0.4 & 0.1 \\
\textit{Stack Cube} & \textbf{0.3} & 0.1 & 0.0 \\
\midrule
\textbf{Avg.} & \textbf{0.467} & 0.317 & 0.117 \\
\bottomrule
\end{tabular*}
\end{table}

As shown, our system achieved moderate success in dynamic tasks like \textit{Pouring} (6/10) and \textit{Sweep Table} (5/10), demonstrating adaptability to non-rigid objects and varying scene layouts. In high-precision tasks like \textit{Stack Cube} (3/10) and \textit{Open Drawer} (3/10), performance was lower than in simulation, highlighting the challenges of depth sensing noise and calibration accuracy in real-world settings. Despite these limitations, our method achieves a much higher average success rate (46.7\%) than DP3 (31.7\%) and DP (11.7\%), validating the strong general applicability and robustness of the proposed dual-stream architecture in physical environments. 

\section{CONCLUSIONS}

We presented ReMAP-DP, a framework that equips generalist robot policies with explicit 3D spatial awareness while preserving pre-trained semantic priors. By re-projecting aggregated multi-view RGB-D data onto standardized planes, our dual-stream architecture systematically overcomes point cloud irregularity and camera dependency. Furthermore, leveraging modality embeddings enables precise, token-level alignment between geometry and semantics for diffusion-based action generation. Empirical results demonstrate ReMAP-DP's superior spatial precision, data efficiency, and robustness across diverse simulated and real-world tasks. In future work, we intend to augment our framework with active viewpoint selection mechanisms and extend this robust multi-modal alignment paradigm to large-scale Vision-Language-Action models.

\bibliographystyle{./IEEEtran} 
\bibliography{./IEEEabrv,./IEEEexample}

\end{document}